%% file: main.tex
\documentclass{article}

\usepackage{arxiv}

\usepackage[utf8]{inputenc} 
\usepackage[T1]{fontenc}    
\usepackage{hyperref}       
\usepackage{url}            
\usepackage{booktabs}       
\usepackage{amsfonts}       
\usepackage{nicefrac}       
\usepackage{microtype}      
\usepackage{lipsum}
\usepackage{graphicx}
\usepackage{multirow}
\usepackage{rotating}
\usepackage{amsmath}
\usepackage{subcaption}
\graphicspath{ {./images/} }

\title{The Influence of Faulty Labels in Data Sets on Human Pose Estimation}

\author{
 Arnold Schwarz \\
  Berliner Hochschule für Technik\\
  Berlin, Germany \\
  \texttt{arnold.schwarz@bht-berlin.de} \\
   \And
 Levente Hernadi \\
  Berliner Hochschule für Technik\\
  Berlin, Germany \\
  \texttt{leventejanko.hernadi@bht-berlin.de} \\
  \And
 Felix Bießmann \\
  Berliner Hochschule für Technik\\
  Berlin, Germany \\
  \texttt{felix.bießmann@bht-berlin.de} \\
  \And
 Kristian Hildebrand\\
  Berliner Hochschule für Technik\\
  Berlin, Germany \\
  \texttt{kristian.hildebrand@bht-berlin.de} \\
}

\begin{document}
\maketitle

\input{0_abstract}

\input{1_intro}

\input{2_related_work}

\input{3_method}
\input{4_experiments}
\input{5_conclusion_and_co}

\section*{Acknowledgments}

This research is funded by the German Research Foundation (DFG) - Project number: 528483508 - FIP 12.

\input{output.bbl}
\input{appendix}

\end{document}

%% file: 0_abstract.tex
\begin{abstract}
In this study we provide empirical evidence demonstrating that the quality of training data impacts model performance in Human Pose Estimation (HPE). Inaccurate labels in widely used data sets, ranging from minor errors to severe mislabeling, can negatively influence learning and distort performance metrics. We perform an in-depth analysis of popular HPE data sets to show the extent and nature of label inaccuracies.
Our findings suggest that accounting for the impact of faulty labels will facilitate the development of more robust and accurate HPE models for a variety of real-world applications. We show improved performance with cleansed data.

\end{abstract}

%% file: 1_intro.tex
\section{Introduction}
\label{sec:intro}

Human Pose Estimation (HPE) has recently been used to analyse and make decisions at sporting events. These analyses of human pose in the centimetre range can be decisive. The uncertainties or the quality of the machine decision making are partly due to the underlying data quality and are usually not communicated to the analyst. 
While HPE has a long history of previous work, current model-based approaches mostly rely on two datasets for training (see~\autoref{fig:dataset_usage_train_bench}). These datasets are labelled for 2D HPE, which forms the basis for 3D HPE in many newer approaches ~\cite{pavllo20193d,shan2022p,zhang2022mixste,Zhan_2022_CVPR,Tian_2023} and is crucial for further applications, including those mentioned above.

\begin{figure}[h]
    \centering
    \includegraphics[width=1\linewidth]{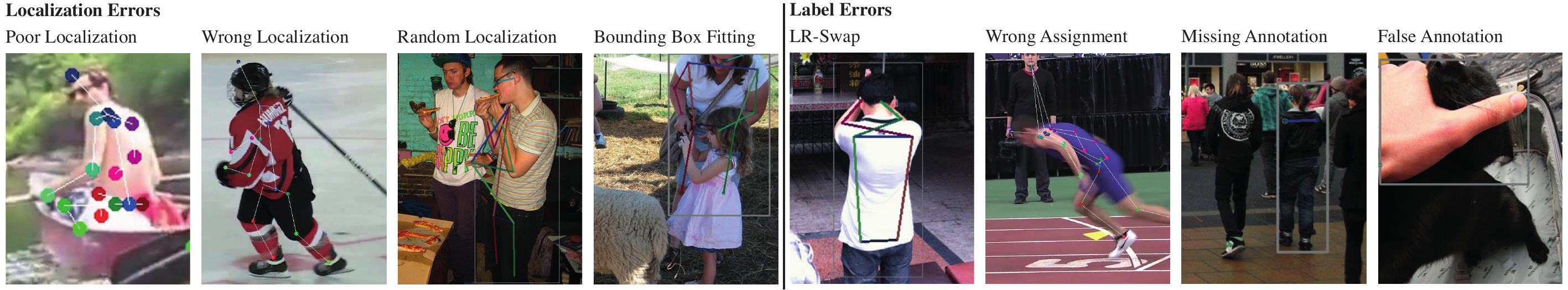}
    \caption{We present an error taxonomy specifically for keypoint label errors in HPE. Left: We categorize and summarize various types of localization errors. Right: We display a range of label errors, highlighting their diversity.}
    \label{fig:taxonomy}
\end{figure}

While such standardized benchmarks are a fundamental prerequisite to scientific progress, recent studies have also highlighted problems with this approach. For one, with the ever increasing pace of innovation in the field of Machine Learning (ML), models achieve very good results in benchmarks quickly and the impact of model improvements become difficult to measure when predictive performance saturates \cite{srivastavaImitationGameQuantifying2023}. This problem is aggravated by the fact that statistical significance of performance improvements in such benchmarks is often difficult to assess: The generalization error on the test set is typically reported as a point estimate, but it can exhibit variances often larger than the differences in predictive performance commonly reported at the top of benchmark leaderboards~\cite{bouthillierAccountingVarianceMachine2021}. Hence improvements on benchmarks that do not account for the uncertainty in the generalization error estimates are sometimes difficult to evaluate~\cite{steinbachMachineLearningStateoftheArt2022}.
Second, data quality issues in the training and test data even in well curated and carefully controlled benchmark data sets~\cite{ronchi2017benchmarking} have been demonstrated to impact both model training and estimates of generalization performance~\cite{northcuttPervasiveLabelErrors2021}.
In this study we argue that the impact of these problems on the advances of the state of the art in HPE have been underrepresented in the literature so far. We provide empirical evidence for data quality problems in erroneous annotation data of commonly used data sets and we demonstrate their impact on model training and generalization performance.
These findings have implications for the interpretation of previously published results. 

Therefore we underline the importance to look at the training and evaluation data to ensure reliability in benchmarks and real-world scenarios. In this work we contribute the following to existing methods and research:
\begin{enumerate}
    \item We provide a comprehensive analysis of labeling errors in commonly used HPE benchmark data sets, developing both a taxonomy of annotation errors and a simple yet effective heuristic to identify faulty and difficult annotations. 
    \item We evaluate the proposed heuristic's impact on data quality in established models leading public HPE benchmarks, discussing the effects of cleaned data and model improvements on popular HPE models.
\end{enumerate}

%% file: 2_related_work.tex
\section{Related Work}

\subsection{Human Pose Estimation}
There have been several surveys looking into the current state of HPE technology \cite{Zheng2023_human_pose_est_survey, Topham2022_survey, Kumar2022_survey, Chen20222d_survey}. HPE is generally divided into whether the model regresses the body keypoints directly as pixel coordinates or whether it uses a probabilistic prediction by outputting keypoint coordinates as heatmaps, which both show notable results \cite{Zheng2023_human_pose_est_survey}. Furthermore, there are two widely used paradigms 2D HPE falls into: the top-down- or the bottom-up approaches. The bottom-up approach initially identifies all visible body keypoints in an image, and subsequently matches these keypoints to the respective individuals. In contrast, the top-down method begins by detecting each person and their bounding boxes, followed by pinpointing the keypoints within each of these boxes.

\subsection{Training and Benchmarking Data sets}
HPE models are typically trained and evaluated on specific data sets, with their performance assessed using a variety of benchmark metrics tailored to each data  set. The de facto standard among these data sets are COCO\footnote{\url{https://cocodataset.org}} (Common Objects in Context)~\cite{Lin2014_COCO} and MPII\footnote{\url{http://human-pose.mpi-inf.mpg.de/}} (Max Planck Institute for Informatics)~\cite{Andriluka14cvpr} due to their size (200.000 and 40.000 labeled poses for COCO and MPII respectively) and their variety in terms of image content. For performance measurement, different metrics are employed based on the data set. For instance, MPII commonly uses a variant of the Percentage of Correct Keypoints (PCK) metric, specifically PCKh@0.5. This metric measures keypoint detection accuracy, considering a keypoint correct if it is within a threshold distance (50\% of the head segment length for PCKh@0.5) from the ground truth. COCO uses the Object Keypoint Similarity (OKS) score, a metric that considers the person's scale and keypoint visibility in the image, and evaluates the similarity between predicted and actual keypoints, allowing for some labeling noise.
Since most models evaluate and train on COCO and MPII we analyze those in more detail in~\autoref{sec:dataset_usage_train_bench}.

\subsection{Advancements in Keypoint Detection}

A notable advancement in pixel-wise regression and keypoint detection accuracy was achieved through HRNet~\cite{sun2019deep}, which addresses information loss challenges by implementing an architecture consisting of parallel subnetworks with varying resolutions. 
This makes HRNet a preferred backbone for state-of-the-art keypoint prediction models and is used throughout in this study. Other advances tackled the issue of accuracy loss introduced by image and annotation pre- and post-processing such as Unbiased Data Processing (UPD)~\cite{huang2020devil} and Distribution-Aware Coordinate Representation (DARK)~\cite{zhang2019distributionaware}. Recent leaderboards on the COCO and MPII test-set often mention the following high performing technologies: Polarized Self-Attention (PSA) \cite{liu2021polarized},  Residual Step Network (RSN)\cite{cai2020_learningdelicatelocalrepr} and ViTPose \cite{xu2022vitpose}. PSA uses the idea of Self-Attention layers \cite{vaswani2023attention} coupled with the concept of polarized filtering.
RSN introduced intra-level feature fusion through dense connections in its \emph{Residual Step Blocks} to refine feature representation at sequential convolution levels and VitPose utilized the popular transformer architecture for the keypoint detection task. In Table~\ref{table:state_of_the_art_method_performance_scores} we report scores for each method as they have been published in their respective work. 

\begin{table}[ht]
\caption{Reported results of state-of-the-art 2D human pose estimation methods for the OKS metric (COCO test-dev) and the PCKh@0.5 (MPII)}
\label{table:state_of_the_art_method_performance_scores}
\centering
\begin{tabular}{lll}
\hline
Method  & AP (COCO) & PCKh@0.5 (MPII)  \\ 
\hline
VitPose+/VitPose-G &\textbf{81.1}\% & \textbf{94.3} \\
HRNet + UDP + PSA     &79.4\%        & -  \\
RSN     &79.2 \%        &93.0\%          \\       
HRNet + DARK    &76.2\%         &90.6 \% \\
HRNet + UDP    &76.5\%         & - \\
\hline
\end{tabular}
\end{table}

\subsection{Data Quality in Machine Learning}
Data quality has been recognized as one of the most important hyperparameters in ML model development. While the majority of models assume stationarity of both the data distribution as well as the label distribution, this assumption is violated in most real world applications. Data errors, or more generally shifts in the data distribution, do occur often and have been studied actively \cite{kimTaxonomyDirtyData}. In the ML community this research is often referred to \textit{covariate shift} if the shift is attributed to the input features \cite{sugiyamaMachineLearningNonStationary2012} and \textit{label shift}, if the shift is associated with the target variable \cite{liptonDetectingCorrectingLabel2018c}. Some sources of noise in training data can have positive, regularizing effects \cite{bishopTrainingNoiseEquivalent1995}. This effect is  leveraged in the augmentation techniques applied in computer vision to render the models invariant w.r.t. data shifts that do not change the semantic properties of an image. Other data quality problems have been demonstrated to have severe negative impact on generalization performance \cite{schelter2021jenga,northcuttPervasiveLabelErrors2021,Hasan2022_NoiseID}. Consequently strategies to detect data quality problems in data sets are being investigated in various application domains \cite{abedjanDetectingDataErrors2016}. A key challenge in this context remains automation \cite{biessmannAutomatedDataValidation2021}. Several approaches were proposed to detect data set shifts \cite{schelterChallengesMachineLearning2018,breckDataValidationMachine2019,rabanserFailingLoudlyEmpirical2018} or label shifts \cite{liptonDetectingCorrectingLabel2018c}. Other approaches aim at generation of realistic errors to improve model robustness \cite{Algan2020_labelnoise,Chong2022_DetectingLE}. 
Our work is inspired by and complements recent findings that demonstrate how simple heuristics can be effective for removing label errors in computer vision benchmark data sets and significantly impact the generalization error estimates \cite{northcuttPervasiveLabelErrors2021}.

\subsection{Label Noise in HPE}
\label{sec:related_work_label_noise_hpe}
In the particular application domain investigated in this study, 2D HPE, label noise has been discussed by several authors and is typically being reported as:

\begin{itemize}
    \item Missing keypoints for visible body parts \cite{Kato2018_HPE_noise_ImprovingMP}.
    \item Localisation errors of keypoints \cite{Johnson2011_HPE_noise, Wan2023_HPE_noise, ronchi2017benchmarking}.
    \item Structure confusion (Left/Right, Arms/Legs) \cite{Johnson2011_HPE_noise}.
    \item Randomly annotated keypoints \cite{Johnson2011_HPE_noise}.
    \item Missing or incomplete occlusion label \cite{Kato2018_HPE_noise_ImprovingMP}.
\end{itemize}

Most approaches focus on making the HPE models more robust in the presence of noisy labels, rather than cleaning data. Johnson et al.\cite{Johnson2011_HPE_noise} model keypoint localization errors produced by human workers as an isotropic Gaussian distribution of vertical and horizontal displacement. Structural errors are assumed to be uniformly distributed across the entire data set. Using a small subset of 'expert' annotated data they define a learning task to steer faulty annotations closer to the 'expert' truth. 

Kato et al.~\cite{Kato2018_HPE_noise_ImprovingMP} adapt the concept of knowledge distillation by using a teacher model to improve insufficient ground truth labels. A student network is then trained using improved annotations to increase its performance. To handle missing, shift, and duplicate noise in point data Wan et al.~\cite{Wan2023_HPE_noise} proposed a new loss function that takes uncertainty about labels into consideration.

Complementing this prior work we propose to investigate in more detail the types of errors and their impact on HPE model training and generalization error estimates. While previous methods accept the errors in the data set, we show their frequencies and suggest a method for detecting those outliers. By data cleansing and showing the effect on training and evaluation, we conclude the problems in effectiveness of existing models trained and evaluated on these data sets, their benchmark in general and its implications on real-world scenarios. We show a more detailed error taxonomy for keypoint label errors in Figure~\ref{fig:taxonomy}.

%% file: 3_method.tex
\section{Method}

We determine and quantify the prevalence and type of errors in the data set using a systematic analysis. This involves a comprehensive examination of various data sets (\ref{sec:dataset_usage_train_bench}), leading to the development of a detailed error taxonomy (\ref{Subsec:dataset_errors}). Utilizing this taxonomy, we then devise a strategy for detecting faulty labels, enabling us to identify and address errors across the entire data set (\ref{subsec:label_noise}-\ref{sec:threshold_calibration}).

\subsection{Data Set Selection}
\label{sec:dataset_usage_train_bench}

\begin{figure}[ht]
    \centering
    \includegraphics[width=0.7\linewidth]{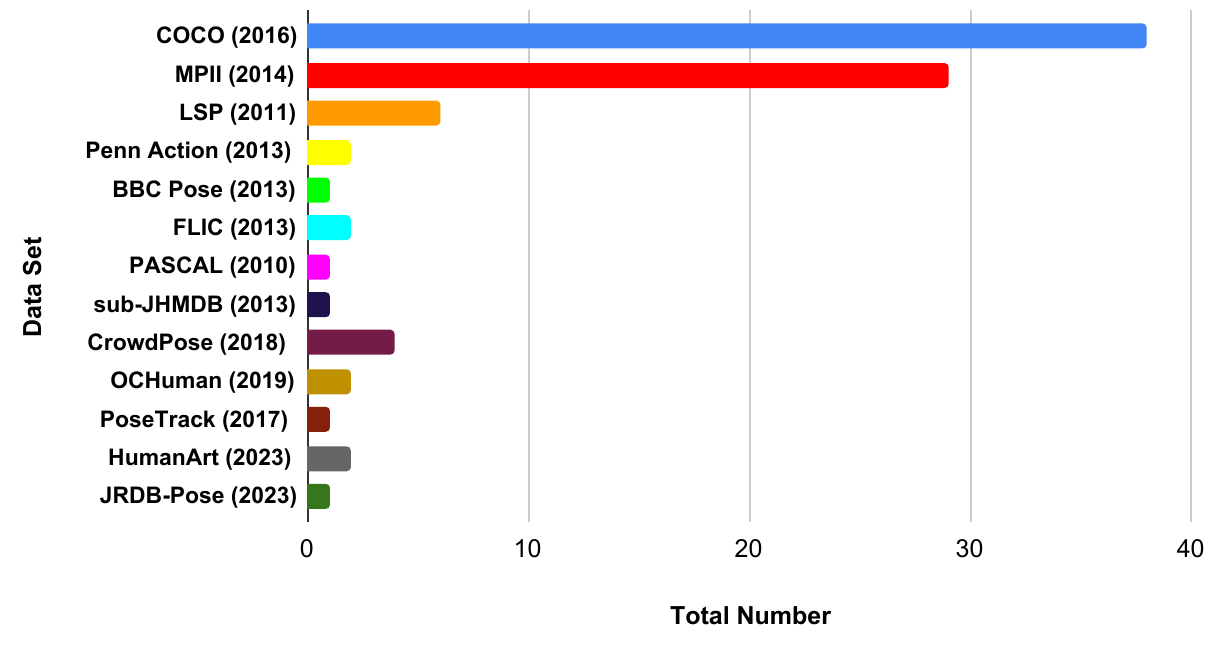}
    \caption{The plot illustrates the frequency of usage of each data set for training or benchmarking purposes. Data was gathered from three HPE surveys, encompassing 40 distinct methods, and additional research for publications after 2023.}
    \label{fig:dataset_usage_train_bench}
\end{figure}

Due to the variety of different training and benchmarking data sets, we decided to reduce our selection to the most relevant data sets currently used for HPE. We evaluated how many unique models and methods used which data set based on data from three surveys \cite{Topham2022_survey, Chen20222d_survey, Zheng2023_human_pose_est_survey} which included 40 different methods from 2018 to 2022. In order to include publications after 2023 we also considered additional research working with online databases. The results can be seen in Figure~\ref{fig:dataset_usage_train_bench}. The outcome of this evaluation indicates that COCO and MPII are the most commonly used data sets for training and testing. Many published papers rely on the training data they provide and on their evaluation tasks to assess the performance of their methods. The ground truth annotation data is often provided by a crowd of human annotators, for example the Amazon Mechanical Turk (AMT) platform\cite{Andriluka14cvpr}. As mentioned earlier by Johnson et al \cite{Johnson2011_HPE_noise}, human crowd work on pose labelling is prone to errors. Some data sets like COCO model margins of human labelling noise into their evaluation metric \cite{ronchi2017benchmarking}, however severe labelling mistakes not accounted by the metric might still influence training and validation results.

\begin{figure}[ht]
    \centering
        \includegraphics[width=0.5\linewidth]{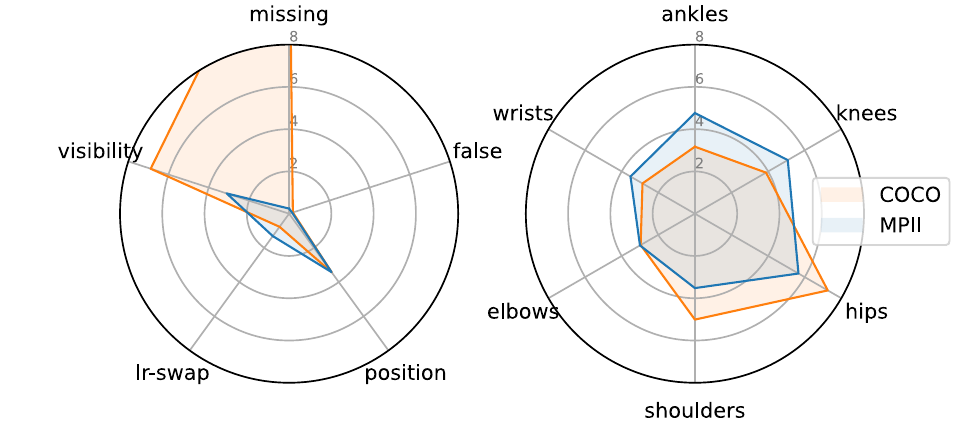}
        \caption{Left: The frequency of occurrence of an error class in our sample. Right: The frequency of error notes for keypoint classes (consider only false annotations, LR-swap and position errors).}
        \label{fig:error_frequencies}
\end{figure}

\subsection{Data Set Errors}
\label{Subsec:dataset_errors}

We define an extended error taxonomy for the data sets as shown in~\autoref{fig:taxonomy} and distinguish between two main error classes, localization errors and labeling errors. For localization errors, we distinguish different subclasses to describe specific types of wrong positioning of keypoints or bounding boxes. These errors are continuous, while labeling errors are discrete and describe different types of missing labels or errors in the data structure.

In order to better annotate the error frequency, we have reduced the error taxonomy (see Figure~\ref{fig:taxonomy}) to 5 classes and summarized them: 
\begin{itemize}
    \item Missing annotation: a keypoint is located on the image plane, is visible or not and is not labeled.
    \item False label: A keypoint is not on the image plane but is labeled.
    \item Incorrect position: The position of the keypoint is clearly incorrect.
    \item Left-right swap: Two keypoints and their assignment are swapped between left and right.
    \item Visibility error: The visibility flag of the key point is set incorrectly.
\end{itemize}
In a next step, we asked three people to label these five error classes in a sample of the MPII validation set (here we use the split from HRNet~\cite{sun2019deep}) and the COCO data set. From the MPII validation set we randomly take 161 out of 4917 annotations and from the COCO validation set we take 163 out of 6352. This corresponds to a confidence level of 99\% and a margin of error of 10\%.

Our analysis reveals that both data sets contain approximately 2\% positional and left-right swap errors, along with false annotations. In these cases, keypoints are incorrectly labeled, resulting in them not actually being present on the image plane (see~\autoref{fig:taxonomy}). As can be seen in~\autoref{fig:error_frequencies}, both data sets show a large error for \emph{visibility flag labels}, especially the COCO data set. Although these errors do not affect current top-down approaches, since they are not considered during training and loss calculation, they can, however, influence bottom-up approaches \cite{CHENG2023109403bottomUp}.

We should also mention that the results between the two data sets are not directly comparable, as our annotators may have been better trained to detect errors in the COCO data set, leading to a seemingly more critical approach to scoring.
We did not additionally evaluate the bounding box fitting in the manual evaluation process. However, a frequency of poor bounding box fitting was noted. If we only look at the raw annotation data, we can already see that in the MPII data set around 3.4\% of the annotated keypoints are outside the ground truth bounding box. For the COCO data set it is around 3.8\%. If the ground truth bounding is used in the evaluation, then it may be the case that the keypoint cannot be estimated at all, as the part is cut off in the top-down process.

\subsection{Automatic Label Noise Detection}\label{subsec:label_noise}

In order to detect data points with faulty labels $y_e$ automatically with high confidence we develop heuristics to estimate $p(y_e|\hat{y},y)$, the probability that a label is wrong given the prediction of a HPE model $\hat{y}\in\mathbb{R}^2$ and the (potentially faulty) annotation $y\in\mathbb{R}^2$. We assume that the majority of data points is correctly annotated and that a HPE model has learned to make accurate predictions. Large deviations between prediction $\hat{y}$ and annotations $y$ hence indicate labeling errors. 
We assume that the large deviations between prediction $\hat{y}_m$ and annotations $y$ are similar for each HPE model $m\in\{1,\dots,M\}$ described in~\autoref{subsec:model_set}.  We use the distance $\delta_m$ between the predictions $\hat{y}_m$ of model $m\in\{1,\dots,M\}$ and the ground truth $y$ 
\begin{align}
\delta_m=\hat{y}_m-y
\end{align}

to create a feature vector $\Delta= [\delta_1, \delta_2, \ldots, \delta_M]$. Our heuristic aggregates the distances of single joints estimates into one score by modeling the distribution of deviations $p(\Delta)$ across all models

\begin{align}
p(y_e|\hat{y},y) = 1 - p(\Delta),
\end{align}

For estimating $p(\Delta)$ we use a well established non-parametric outlier detection method, an \textit{Isolation Forest} \cite{4781136} as implemented in PyOD \cite{zhao2019pyod}. We emphasize that the choice of the outlier detection method is not the key factor for the label noise detection proposed in this work. The relevant functionality is a non-parametric density estimator that approximates the distribution of deviations $p(\Delta)$ given the multivariate feature vectors $\Delta$. Indeed we find most other methods commonly used for outlier detection to work equally well.

\subsection{Evaluation Models}
\label{subsec:model_set}

We select five different top-down approach models using the MMPose\cite{mmpose2020} library to create and evaluate our label noise detection.  
MMPose supports various 2D human pose estimation model architectures and data sets. Therefore, it enabled us to keep evaluation standardized and fair by using the configuration files for each model listed in \autoref{table:cleaning-result-table-coco} and \autoref{table:results_all}. To simplify the reproduction of results we used pre-trained model checkpoints provided by MMPose to generate the predictions for the outlier detection. The model performances on the original validation set for MPII and COCO, reported in this paper, are all consistent with scores reported by MMPose\footnote{\url{https://mmpose.readthedocs.io/en/latest/overview.html}}. Deviations in the COCO results compared to the results listed online are due to the fact that ground truth bounding boxes were used for the COCO evaluation, to avoid an influence of the bounding box estimation errors on the metric evaluation.

We ensured models were comparable in size and performance. Model selection was mostly kept consistent between the COCO and MPII data set. However, for models pre-trained on the COCO data set there was no Hourglass model available that shared the same input size (256x192) with the other models. For COCO, we therefore, exchanged Hourglass for ResNeSt to maintain higher unity between models.

 \subsection{Per Keypoint Distance for MPII and COCO}
 \label{subsec:keypoint_distance}

Our non-parametric outlier detection requires the distance errors per joint. In order to extract this information we made the following modifications to the MPII and COCO evaluation pipelines. For MPII we modified the MMPose evaluation script so that per keypoint prediction to ground truth distance were saved in addition to the mean score values. 
For the COCO data set, we used a modified version of the original COCO evaluation code\footnote{\url{https://github.com/cocodataset/cocoapi}}. For each model prediction we saved the OKS metric per joint and pose as well as the distance for prediction to ground truth. Note that we did not use the 'raw' distance between prediction and ground truth but the modified version that takes the OKS $\sigma$ values and the object scale into consideration.
We want to note here that a person in an image can have multiple predictions, which, during the OKS calculation, are filtered using different IoU (intersection over union) thresholds. A lower threshold comes with a higher recall, which means more poses are found. For the heuristic we determined to use the 'loose' 0.5 IoU threshold to extract distance and OKS score per joint. 

\subsection{Calibrating the Outlier Threshold} 
\label{sec:threshold_calibration}
\begin{figure}[ht]
    \begin{subfigure}[t]{0.49\linewidth}
        \includegraphics[width=1\linewidth]{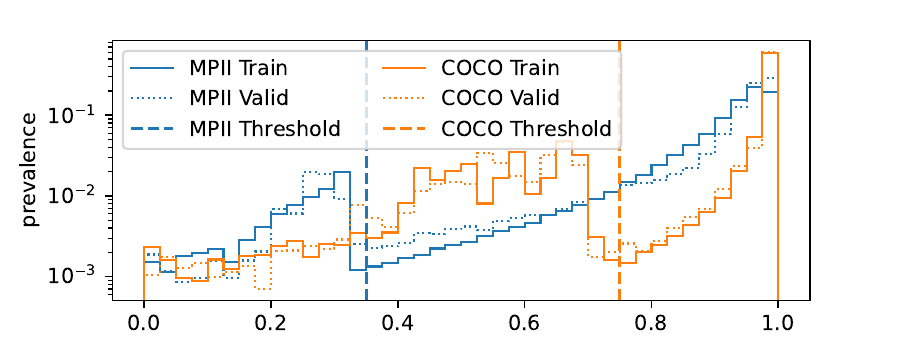}
        \caption{The score distribution of all possible keypoints, including the keypoints without annotation.}
        \label{fig:outlier_scores_w_invis}
    \end{subfigure}
    \hfill
    \begin{subfigure}[t]{0.49\linewidth}
        \includegraphics[width=1\linewidth]{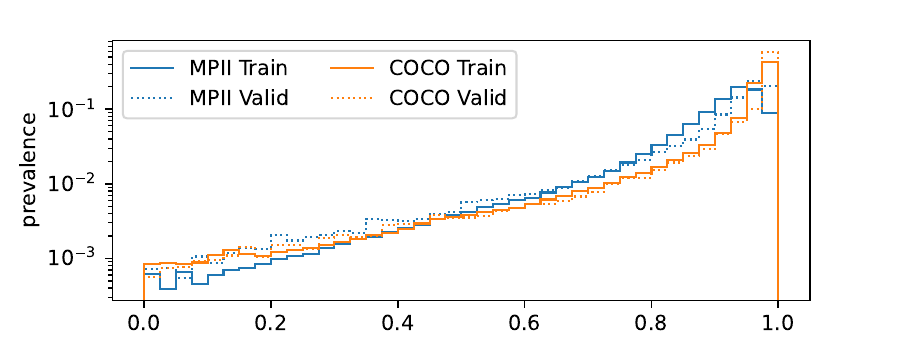}
        \caption{The outlier scores only for the keypoints for which an annotation exists.}
        \label{fig:outlier_scores_wo_invis}
    \end{subfigure}
    \caption{We determine the threshold for faulty annotations using the histograms of outlier scores for poses on images with (a) and without (b) annotations. Including keypoints without annotations, for which predictions are assumed to be less reliable, indicate a clear second mode in the outlier score distribution (compare a vs b). We set the threshold for faulty annotations to exclude all annotations that are as faulty (according to our heuristic) as keypoint predictions without annotations.}
    \label{fig:outlier_scores}
\end{figure}

The outlier score threshold for discarding presumably faulty annotations was calibrated as follows.
We assume that model predictions are less reliable for keypoints for which there is no annotation, for instance because the body part was not on the image plane. HPE models will produce predictions for those poses, but usually these keypoints are not included in the evaluation of HPE models. Here we included these keypoints and computed outlier scores for all keypoints. In \autoref{fig:outlier_scores}a   we show the distribution of all outlier scores for poses with and without keypoint annotations. In \autoref{fig:outlier_scores}b we show the outlier score distribution only for keypoints that do have an annotation. We assume that the erroneous annotations have similar characteristics to the non-annotated keypoints. Keypoints witout annotations form a distinct mode of the outlier score distribution in \autoref{fig:outlier_scores}a. We set the threshold for each data set individually such that outlier scores smaller than the mode of keypoints without annotations are detected as outliers. For the COCO data set the threshold was thus set to 0.75 and for MPII the threshold was set to 0.35. 

%% file: 4_experiments.tex
\section{Experiments and Evaluation}

We conduct a series of experiments and analysis to evaluate the effectiveness of our heuristics. We investigate how refining the validation data set influences model outcomes and assess the effects of various models on the enhanced, label-noise-free data sets. Our comparison includes examining the effect of faulty labels on evaluation metrics for COCO and MPII. We address distinguishing ‘hard’ and ‘faulty’ cases by contrasting manually hand-cleaned data sets (Sec.~\ref{subsection_impact_cleaning_validation_data}) with sets created by our heuristics. Additionally, we examine the influence of annotation errors on the the heatmaps (Sec.~\ref{impact_jitter}). 

\subsection{Label Error Detection Evaluation}
\label{subsec:label_error_detection_eva}

\paragraph{Heuristics evaluation on training set:} To evaluate the performance of the heuristic on the training data sets, we instructed two annotators to label 100 of the 4416 poses in the COCO training set and 100 of the 1102 poses in the MPII training data that were detected by our heuristic.
For the COCO data our heuristic detects faulty labels with an average precision of about 36\% and an average recognition rate of 31\%.
For the MPII data set, the heuristic detects faulty annotations with an average precision of about 41\% and an average recognition rate of 25\%.

\paragraph{Heuristics evaluation on validation set:} 
In order to evaluate the quality of our heuristic to detect faulty annotations in the validation data we instructed three annotators to identify faulty annotations manually on the validation data flagged as faulty by our heuristic. On validation data our heuristic detects faulty annotations with  an average precision of about 16\% and an average recall of 22\% for the MPII validation data set and an average precision of about 15\% and an average recall of 33\% for the COCO validation data set. 

We assume that there are several reasons why our heuristics' performance differs between the training set and the validation set. The models have much better results on the training set, than on the validation set, so separating between faulty and non-faulty keypoints should become easier. Also the annotators have different opinions on keypoint errors, concentration levels and set their focus on different keypoints/body parts. Moreover, the labeling task becomes tedious. Bazarevsky et al.~\cite{DBLP:journals/corr/abs-2006-10204} show that when they let two annotators relabel a data set, they achieve an average PCK@0.2 of 97.2.  This raises questions on the annotation performance in HPE tasks. 

\subsection{Impact of Cleaning Validation Data}
\label{subsection_impact_cleaning_validation_data}

The results in the previous section demonstrate that the proposed heuristic reliably detects faulty labels. To evaluate the impact of erroneous labels on the HPE evaluation metrics  in commonly used benchmarks, we first investigate the impact of cleaning the  validation data sets. We list the results for the COCO data in \autoref{table:cleaning-result-table-coco} and the results for the MPII data in \autoref{table:results_all}. For all metrics we compare results on the original validation data (RAW), the validation data cleaned automatically with our proposed heuristic, here referred to as  \emph{auto cleaned} and the impact of a partial cleanup by a human annotator (HC). We used the annotations from the error frequency analysis (see \autoref{Subsec:dataset_errors}) to clean the data for the HC set. 

\begin{table}[h]
\footnotesize
\centering
\caption{Impact of cleaning on HPE metrics computed on COCO data. Compared are metrics obtained on the original data set (RAW), the manually cleaned set (HC) and the automatically cleaned data set (AC). Results obtained when cleaning training data (TC) are listed in the bottom row. Predictive performance is improved when training data is cleaned with the proposed heuristic. Cleaning validation data leads to slight improvements, too. The input size for all models listed was 256x192}
\label{table:cleaning-result-table-coco}
\begin{tabular}{lcccccc}
 & \multicolumn{3}{c}{AP} & \multicolumn{3}{c}{AR} \\
 & RAW & HC & AC & RAW & HC & AC \\
 \toprule
ResNet50 \cite{he2015deep} & 73.6 & 73.6 & 74.5 & 76.6 & 76.7 & 77.4 \\
ResNeSt \cite{zhang2020resnest} & 73.8 & 73.8 & 74.7 & 76.8 & 76.8 & 77.6 \\
SE-ResNet50 \cite{hu2018squeeze}  & 74.5 & 74.7 & 75.5 & 77.7 & 77.8 & 77.5 \\
SCNet50 \cite{liu2020improving} & 74.6 & 74.6 & 75.6 & 77.7 & 77.7 & 78.5 \\
HRNet\_w32 \cite{sun2019deep} & 76.6 & 76.7 & 77.5 & 79.3 & 79.4 & 80.2 \\
\toprule
HRNet\_w32 \cite{sun2019deep} TC & \textbf{76.7} & \textbf{76.8} & \textbf{77.6} & \textbf{79.4} & \textbf{79.5} & \textbf{80.2} \\
\bottomrule
\end{tabular}
\end{table}

\begin{table}[h]
\footnotesize
\centering
\caption{Impact of cleaning on HPE metrics computed on MPII data set. Compared are metrics obtained on the original validation data set (RAW), the manually cleaned set (HC) and the automatically cleaned data set (AC). Results obtained when cleaning training data (TC) are listed in the bottom rows. Especially when removing faulty labels from the training data we observe improvements in predictive performance. The input size for all models was kept consistent to 256x256}
\label{table:results_all}
\begin{tabular}{lcccccc}
 & \multicolumn{3}{c}{PCKh@0.5} & \multicolumn{3}{c}{PCKh@0.1} \\
 & RAW & HC & AC & RAW & HC & AC \\
 \toprule
ResNet50 \cite{he2015deep} & 88.2 & 93.6 & 94.4 & 28.6 & 66.7 & 67.3 \\
SE-ResNet50 \cite{zhang2020resnest}   & 88.4 & 93.8 & 94.5 & 29.2 & 66.8 & 67.4 \\
SCNet50 \cite{liu2020improving}  & 88.8 & 93.9 & 94.7 & 29.0 & 67.3 & 67.9 \\
Hourglass52 \cite{newell2016stacked} & 88.9 & 94.0 & 94.7 & 31.7 & 68.2 & 68.8 \\
HRNet\_w32 \cite{sun2019deep} & 90.0 & 94.5 & 95.2 & 33.4 & \textbf{69.7} & \textbf{70.3} \\
\toprule
HRNet\_w32 \cite{sun2019deep} TC &\textbf{90.4} &\textbf{94.7} &\textbf{95.3} & \textbf{33.5} & 69.6 & 70.2 \\
\bottomrule
\end{tabular}
\end{table}

For the MPII validation data set, we discard 469(1.1\%) for auto clean and 161(0.4\%) for hand clean and for COCO auto clean 377(0.6\%) and hand clean 185(0.3\%) keypoint annotations. 

Our results demonstrate that discarding faulty annotations from the evaluation data improves metrics across the board slightly. In some cases the improvements are substantial. For the models trained with the MPII data set, we can see (see additional material) a significant improvement in the results for the Hips, Knees and Ankles. These body parts  seem to be more affected by errors in our studies (\autoref{fig:error_frequencies}). We can also achieve better results for the COCO data set and the benchmark metric with cleaned validation data sets. This time, however, the improvements are significantly smaller. This is especially true for the hand-adjusted validation data set. The modest improvements on COCO could be attributed to the use of a more stringent metric compared to PCKh and we cleaned a smaller subset for COCO. Nonetheless, we can also see a change in the leaderborad order for the SE-ResNet50 and SCNet50 model for the manually cleaned part.

Also, we observe a significant decrease in PCKh@0.5 score variance for the five models on the respective adjusted validation data set. For the raw data set variance was 0.38, which shrunk to 0.09 for the manually cleaned set and 0.08 for the automatically cleaned set. This suggests that the models for the MPII data set are likely to perform similarly well. No such significant decrease is observed for the COCO data set.

\begin{figure}[h]
    \centering
    \begin{subfigure}[t]{0.49\linewidth}
        \includegraphics[width=1\linewidth]{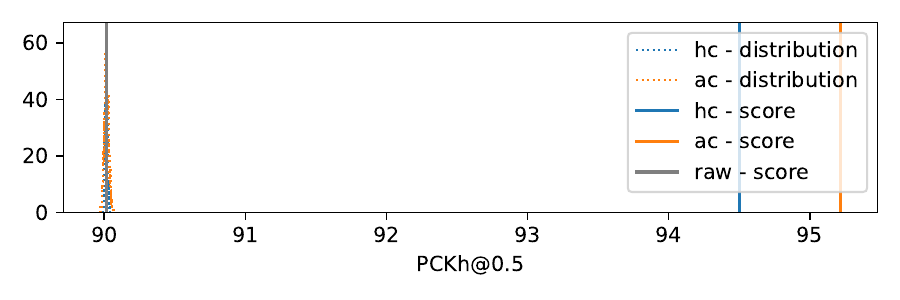}
        \caption{PCKh@0.5 (MPII) - We achieve $509.05\sigma$ for the manually cleaned sample and $325.36\sigma$ for the automatically cleaned sample. Both results therefore show a significant influence on the metrics.
        }
        \label{fig:pckh_dist}
    \end{subfigure}
    \hfill
    \begin{subfigure}[t]{0.49\linewidth}
        \includegraphics[width=1\linewidth]{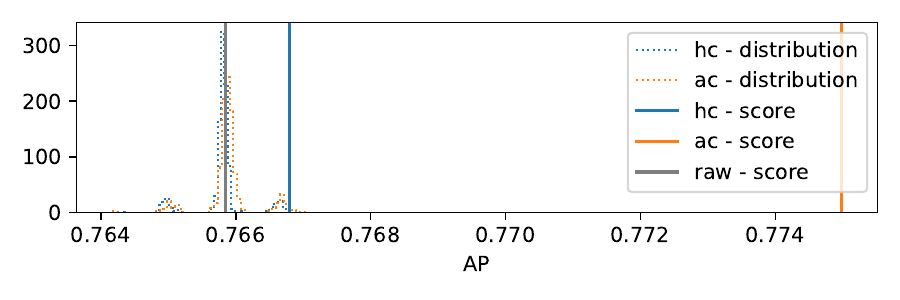}
        \caption{mAP (COCO) - We achieve $2.97\sigma$ for the manually cleaned sample and $25.74\sigma$ for the automatically cleaned sample. Both results therefore show a significant influence on the metrics.
        }
        \label{fig:mAP_dist}
    \end{subfigure}
    \caption{Comparing the accuracy impact of randomly removed keypoints across 1000 repetitions and in comparision hand-cleaned and auto-cleaned data sets.}
    \label{fig:metrics_influence}
\end{figure}

\paragraph{Influence of omitting key points on accuracy:}
To measure the influence of cleaned validation data we illustrate the impact of excluding keypoints from the evaluation on metrics (PCKh and mAP) through HRNet and the corresponding validation set, as depicted in Figure \ref{fig:metrics_influence}. We randomly discarded the same number of keypoints as in the manual and automatic cleanup for the respective data sets and repeated this 1000 times. We can see a distribution of the influence of removing the random keypoints. The mean of each distribution reflects the original score (gray lines - RAW-score- in the \autoref{fig:metrics_influence}). We can also see that we achieve a significant score improvement with the hand-cleaned sets (HC) and the automatically cleaned sets (AC) for the MPII and COCO data set. 

\subsection{Impact of Cleaning Training Data}

Extending previous studies \cite{northcuttPervasiveLabelErrors2021} that only investigated the influence of cleaning on evaluation data, as we did in the previous section, we also investigated the impact of cleaning on the training data.

As shown in \autoref{table:cleaning-result-table-coco} (COCO) and \autoref{table:results_all} (MPII) we observe improved predictive performance when re-training the HRNet model on cleaned data.\\
This effect can be seen for the model trained on cleaned MPII data for almost all body parts except the hips. In the error frequency analysis reported above, annotations for hip keypoints were often affected by labeling errors. Furthermore, hip joints have been reported before by Chen et al.~\cite{chen2018cascaded} as "hard" to predict keypoints for models since their appearance is not always structurally obvious and often in need of more context information. We assume that similar difficulties exist for human annotators to detect hip joints correctly and therefore ground truth positions scatter a lot. 
Nevertheless, the overall result shows that faulty training data influences model performances.\\
Comparing the improvements between the two data sets COCO and MPII we find that improvements for MPII are larger. We assume that this is due to the larger training set of the COCO data set and the higher complexity in COCO. 

\subsection{Impact of Hard Poses}

One explanation for the improvements in evaluation metrics after cleaning the data sets with the proposed heuristic could be that we are not (only) removing faulty annotations -- but just those annotations for poses that are difficult. We investigated this hypothesis with  additional experiments. 
Therefore, we asked three annotators to categorize images into the following three categories:
\begin{itemize}
    \item easy: A pose is relatively easy to annotate without any further assumptions
    \item hard: A pose is time-consuming to annotate or assumptions have to be made
    \item impossible: No pose can be credibly annotated.
\end{itemize}
For the annotations of the COCO data set, we draw the ground truth bounding box on the images. For the annotations of the MPII data set, we draw the center on the image and a quadratic bounding box based on the scale information in the data.
The task for the annotators is to consider only the body parts within the bounding boxes and on the image plane.

\begin{figure}[h]
    \centering
    \begin{subfigure}[t]{0.49\linewidth}
        \includegraphics[width=1\linewidth]{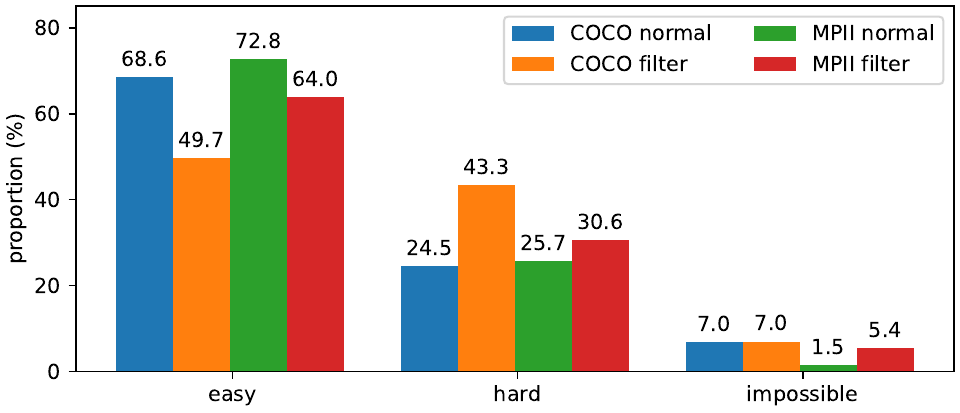}
        \caption{Comparison of the evaluation of the annotators between the sample set of the respective normal data sets and the respective sets discarded by our heuristics.}
        \label{fig:normal_vs_filter}
    \end{subfigure}
    \hfill
    \begin{subfigure}[t]{0.49\linewidth}
        \includegraphics[width=1\linewidth]{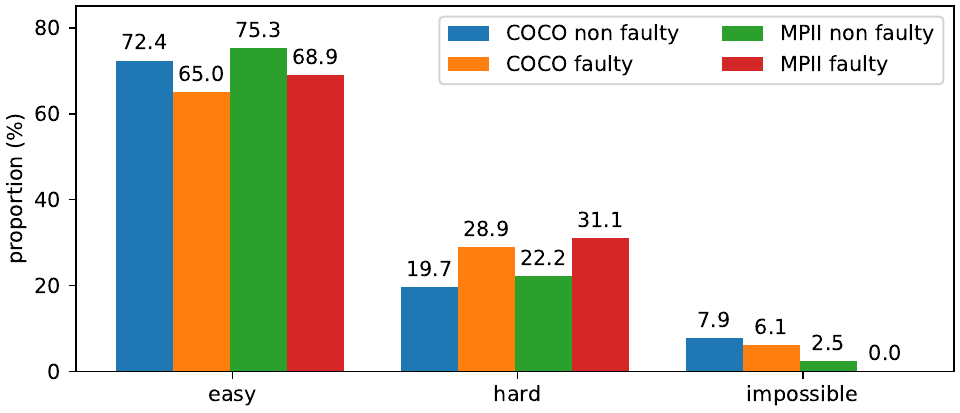}
        \caption{Comparison of the evaluation of the annotators for the normal data set between manually discarded and non-discarded keypoints.}
        \label{fig:non_faulty_vs_faulty}
    \end{subfigure}
    \caption{Evaluation of the poses of the different sets by three annotators in the categories easy, hard and impossible.}
    \label{fig:hard_vs_faulty}
\end{figure}

The evaluation shows that our heuristic indeed has the tendency to discard hard poses (see \autoref{fig:normal_vs_filter}), but we can also see that hard poses are generally more prone to incorrect annotations, as shown in the \autoref{fig:non_faulty_vs_faulty}. Consequently, our heuristic tends to exclude more challenging poses, which happen to also be incorrectly annotated in many cases. Moreover, the evaluation indicates that it also omits annotations that are classified easy by human annotators, suggesting that poses easy for humans might be difficult for models.
As the results of the re-training on the automatically cleaned training sets
in the \autoref{table:cleaning-result-table-coco} and \autoref{table:results_all} show, both models slightly increase their performance on all relevant metrics. 
Even though our heuristic also partially penalizes hard poses, the effect on the models are small (0.1\%). Assuming that the model does not estimate easier poses better now than before, the models do not seem to benefit from hard poses during training.

\subsection{Impact of Annotation Jitter}
\label{impact_jitter}

Current approaches utilize heatmaps for the prediction of keypoints. The accuracy correlates with the variance of these heatmaps; lower variance implies higher precision, while higher variance decreases accuracy which is influenced by several factors. Our findings indicate that the variance inherent in heatmaps stems not only from the model and its hyperparameter but is also significantly influenced by the data quality. This variability is introduced by human annotators who tend to place keypoints slightly differently, an effect considered in the creation of the OKS (Object Keypoint Similarity) score.
To measure the impact of this annotation jitter, we employed the Human3.6M data set~\cite{h36m_pami, IonescuSminchisescu11}, which utilizes ground truth data generated by marker-based motion capture, thereby ensuring minimal annotation jitter. We trained an HRNet model on a subset of the Human3.6M data set and introduced annotation jitter by adding a random normal distribution to the ground truth data. The perturbations were set at $\sigma$ levels of 0.5\%, 1\%, and 2\% of the bounding box diagonal.

\begin{figure}[ht]
    \centering
    \includegraphics[width=0.75\linewidth]{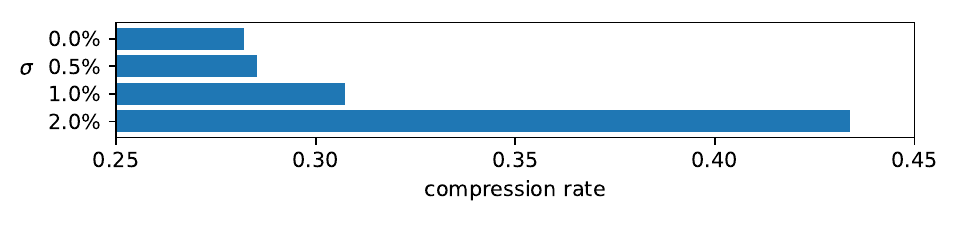}
    \caption{Comparison between annotation jitter and compression ratio using the Human3.6M dataset and HRNet. $\sigma$ is the percentage of the diagonal of the bounding box used for the random normal distribution added to the ground truth data.}
    \label{fig:jitter_comp}
\end{figure}

We evaluate the 4 models based on the validation set and compress the results of the models to determine a compression ratio. We use the compression ratio as an indicator of how noisy the output heatmaps are. As shown in ~\autoref{fig:jitter_comp}, the compression ratio increases with the $\sigma$. This shows that more jitter produces noisier heatmaps. This is particularly relevant as current research~\cite{zhang2019distributionaware,SimCC} focuses on the quality and noise of the heatmaps without investigating the reasons for this.

%% file: 5_conclusion_and_co.tex
\section{Conclusion}
We show that the two most used data sets in the field of human pose estimation contain a variety of faulty annotations. These erroneous annotations have an impact on model training and evaluation. This should motivate further efforts to reach a better understanding of the data sets, which are commonly used in public benchmarks, and that ultimately drive scientific progress. One contribution to achieve this could be a more careful documentation of both data set creation \cite{gebruDatasheetsDatasets2021} and model development \cite{mitchellModelCardsModel2019}. Our results suggest that there is a need to improve model evaluation guidelines by creating more sophisticated testing sets, which also account for data quality in benchmark tasks. 
Only when we know the flaws in the data we use can we truly interpret what our models are doing and where improvements can be made. Specifically for the MPII data set and its latest leaderboard scores, we can assume that we have reached a level in this benchmark, where we can no longer achieve significantly better results. The COCO data set on the other hand, with its more robust metrics leaves more room for further improvement. Furthermore, we show that the performance of HPE models does not only depend on hard cases of poses.

%% file: appendix.tex
\newpage
\appendix

\section{Appendix}

\begin{table}[h]
\centering
\caption{Comparison of the results on the COCO validation data set when evaluated on the original data set (RAW), the manually cleaned set (HC) and the automatically cleaned data set (AC). Results obtained when cleaning training data
(TC) are listed in the right column.}
\label{table:cleaning-result-table-coco-expand}
\begin{tabular}{llccccc|c}
 &  & ResNet50 & ResNeSt & SE-ResNet50 & SCNet50 & HRNet\_w32 & HRNet TC\\
 \toprule
\multirow[c]{3}{*}{AP} & RAW & 73.6 & 73.8 & 74.5 & 74.6 & 76.6 & \textbf{76.7} \\
 & HC & 73.6 & 73.8 & 74.7 & 74.6 & 76.7 & \textbf{76.8} \\
 & AC & 74.5 & 74.7 & 75.5 & 75.6 & 77.5 & \textbf{77.6} \\
\multirow[c]{3}{*}{AP .5} & RAW & 92.5 & 92.5 & 92.6 & 92.6 & \textbf{93.6} & 93.6 \\
 & HC & 92.5 & 92.5 & 92.6 & 92.6 & \textbf{93.6} & 93.6 \\
 & AC & 92.5 & 92.5 & 92.6 & 92.6 & \textbf{93.6} & 93.6 \\
\multirow[c]{3}{*}{AP .75} & RAW & 81.4 & 82.3 & 82.5 & 82.5 & 83.7 & \textbf{84.7} \\
 & HC & 81.4 & 82.3 & 82.5 & 82.6 & 83.7 & \textbf{84.7} \\
 & AC & 82.6 & 83.6 & 83.8 & 83.9 & 85.1 & \textbf{86.0} \\
\multirow[c]{3}{*}{AP (M)} & RAW & 70.7 & 70.6 & 71.9 & 72.1 & 73.7 & \textbf{74.0} \\
 & HC & 70.8 & 70.6 & 72.0 & 72.3 & 73.8 & \textbf{74.1} \\
 & AC & 71.6 & 71.8 & 73.1 & 73.3 & 74.7 & \textbf{75.0} \\
\multirow[c]{3}{*}{AP (L)} & RAW & 78.2 & 78.5 & 78.8 & 78.9 &\textbf{81.2} & 81.0 \\
 & HC & 78.3 & 78.5 & 78.8 & 78.9  & \textbf{81.2} & 81.1 \\
 & AC & 78.8 & 79.0 & 79.6 & 79.5  & 81.8 & \textbf{81.9} \\
\multirow[c]{3}{*}{AR} & RAW & 76.6 & 76.8 & 77.7 & 77.7  & 79.3 & \textbf{79.4} \\
 & HC & 76.7 & 76.8 & 77.8 & 77.7 & 79.4 & \textbf{79.5} \\
 & AC & 77.4 & 77.6 & 78.5 & 78.5 & 80.2 & \textbf{80.2} \\
\multirow[c]{3}{*}{AR .5} & RAW & 93.6 & 93.4 & 94.0 & 93.8 & 94.3 & \textbf{94.4} \\
& HC & 93.6 & 93.4 & 94.0 & 93.8 & 94.3 & \textbf{94.4} \\
& AC & 93.5 & 93.4 & 94.0 & 93.8 & 94.3 & \textbf{94.5} \\
\multirow[c]{3}{*}{AR .75} & RAW & 83.4 & 84.0 & 84.7 & 84.4 & 85.5 & \textbf{86.1} \\
 & HC & 83.4 & 84.1 & 84.7 & 84.5 & 85.5 & \textbf{86.1} \\
 & AC & 84.6 & 85.2 & 85.8 & 85.8 & 86.8 & \textbf{87.2} \\
\multirow[c]{3}{*}{AR (M)} & RAW & 73.4 & 73.3 & 74.7 & 74.7 & 76.1 & \textbf{76.4} \\
 & HC & 73.4 & 73.4 & 74.7 & 74.7 & 76.2 & \textbf{76.5} \\
 & AC & 74.4 & 74.3 & 75.6 & 75.6 & 77.0 & \textbf{77.3} \\
\multirow[c]{3}{*}{AR (L)} & RAW & 81.5 & 82.0 & 82.4 & 82.3 & \textbf{84.3} & 84.2 \\
 & HC & 81.6 & 82.1 & 82.4 & 82.3 & \textbf{84.3} & 84.2 \\
 & AC & 82.2 & 82.6 & 83.0 & 82.9 & \textbf{85.0} & 84.8 \\
 \bottomrule
\end{tabular}
\end{table}

\begin{sidewaystable}
\centering
\caption{\label{cleaning-result-table}Comparison of the results for the MPII data set of publicly available HPE models for individual body parts when evaluated on the original validation data set (RAW), the manually cleaned set (HC) and the automatically cleaned data set (AC) and the results of the model trained on cleaned data (TC). The head part was not considered any further, as the change there was less than 1\% on average.}
\begin{tabular}{lcccccccccccccccccc}
 & \multicolumn{3}{c}{Shoulder} & \multicolumn{3}{c}{Elbow} & \multicolumn{3}{c}{Wrist} & \multicolumn{3}{c}{Hip} & \multicolumn{3}{c}{Knee} & \multicolumn{3}{c}{Ankle} \\
method & RAW & HC & AC & RAW & HC & AC & RAW & HC & AC & RAW 
& HC & AC & RAW & HC & AC & RAW & HC & AC \\
 \toprule
ResNet50    & 95.3 & 95.9 & 96.3 & 88.7 & 93.5 & 94.5 & 83.3 & 91.6 & 93.0 & 87.4 & 92.98 & 93.4 & 83.5 & 92.09 & 92.8 & 78.9 & 90.75 & 92.3 \\
SE-ResNet50 & 95.3 & 96.0 & 96.4 & 88.6 & 93.6 & 94.4 & 83.9 & 91.3 & 92.7 & 87.1 & 93.05 & 93.5 & 83.6 & 92.48 & 93.3 & 80.4 & 91.65 & 93.1 \\
SCNet50     & 95.4 & 96.0 & 96.3 & 88.6 & 93.6 & 94.5 & 84.0 & 91.8 & 93.1 & 88.1 & 93.24 & 93.7 & 84.8 & 92.82 & 93.7 & 80.6 & 91.39 & 93.1 \\
Hourglass52 & 95.4 & 96.0 & 96.3 & 89.3 & 94.0 & 94.8 & 84.0 & 91.8 & 93.1 & 87.9 & 92.94 & 93.3 & 84.5 & 93.09 & 93.8 & 80.6 & 91.49 & 92.9 \\
HRNet\_w32  & 95.8 & 96.4 & 96.7 & 90.4 & 94.4 & 95.1 & 85.8 & 92.6 & \textbf{93.9} & 89.2 & \textbf{93.81} & \textbf{94.2} & 86.3 & 93.63 & \textbf{94.4} & 82.3 & 92.34 & 93.8 \\
\toprule
HRNet\_w32 TC   & \textbf{96.2} & \textbf{96.5} & \textbf{96.8} & \textbf{90.8} & \textbf{94.8} & \textbf{95.5} & \textbf{86.1} & \textbf{92.8} & \textbf{93.9} & \textbf{89.4} & 93.67 & 94.0 & \textbf{86.7} & \textbf{93.71} & \textbf{94.4} & \textbf{82.9} & \textbf{92.81} & \textbf{94.1} \\
\bottomrule
\end{tabular}
\end{sidewaystable}